\documentclass[letterpaper, 10 pt, conference]{ieeeconf}  % Comment this line out if you need a4paper

\IEEEoverridecommandlockouts                              % This command is only needed if 
\usepackage[utf8]{inputenc} % allow utf-8 input
\usepackage[T1]{fontenc}    % use 8-bit T1 fonts
\usepackage{url}            % simple URL typesetting
\usepackage{booktabs}       % professional-quality tables
\usepackage{amsfonts}       % blackboard math symbols
\usepackage{nicefrac}       % compact symbols for 1/2, etc.
\usepackage{microtype}      % microtypography
\usepackage{xcolor}         % colors

\usepackage{amsmath}
\usepackage{amssymb}
\usepackage{wrapfig}
\usepackage{subfig}
\usepackage{caption}
\usepackage{color}
\usepackage{algorithm}
\usepackage{algorithmic}
\usepackage{graphicx}

\usepackage{pgfplots}
\usepackage{tikz}
\usepackage{tikz-3dplot}

\usepackage{ragged2e}
\usepackage{fp}

\usepackage{ifthen}

\usepackage{tikz}
\usetikzlibrary{shapes.geometric, arrows, calc,automata,positioning,decorations,decorations.text}

\usetikzlibrary{mindmap,snakes}

\usepackage{multirow}

\usepackage{algorithm}
\usepackage{algorithmic}
\usepackage{graphicx}
\usepackage[english]{babel}
\usepackage{array}
\usepackage[export]{adjustbox}

\usepackage{nicefrac}

\usepackage{xcolor}

\usepackage{colortbl}
\usepackage{makecell}
\usepackage{pifont}

\graphicspath{{images/}}

\usepackage[colorlinks=false,
    linkcolor=red,
    linkbordercolor=red,
    urlcolor=red!90!blue,
    urlbordercolor=white
    ]{hyperref}

\title{\LARGE \bf DeepMI: A Mutual Information Based Framework For Unsupervised Deep Learning of Tasks}

\author{Ashish~Kumar$^{\dagger}$, L. Behera$^{\dagger}$, \textit{Senior Member IEEE}
 \\
\thanks{$^{\dagger}$Department of Electrical Engineering, Indian Institute of Technology, Kanpur
        {\tt\small \{krashish,lbehera\}@iitk.ac.in}}%
}

\begin{document}

\maketitle
\thispagestyle{empty}
\pagestyle{empty}

\begin{justify}

\begin{abstract}
In this work, we propose an information theory based framework ``DeepMI'' to train deep neural networks (DNN) using  Mutual Information ($\mathcal{MI}$). The DeepMI framework is especially targeted but not limited to the learning of real world tasks in an unsupervised manner. The primary motivation behind this work is the limitation of the traditional loss functions for unsupervised learning of a given task. Directly using $\mathcal{MI}$ for the training purpose is quite challenging to deal with because of its unbounded above nature. Hence, we develop an alternative linearized representation of $\mathcal{MI}$ as a part of the framework. Contributions of this paper are three fold: \textit{i}) investigation of  $\mathcal{MI}$ to train deep neural networks, \textit{ii}) novel loss function $\mathcal{L}_{LMI}$, and \textit{iii}) a fuzzy logic based end-to-end differentiable pipeline to integrate DeepMI into deep learning framework. Due to the unavailabilty of a standard benchmark, we carefully design the experimental analysis and select three different tasks for the experimental study. We demonstrate that $\mathcal{L}_{LMI}$ alone provides better gradients to achieve a neural network better performance over the popular loss functions, also in the cases when multiple loss functions are used for a given task.
%
%In this work, we present a fully differentiable information theoretic based formulation DeepMI to train deep neural networks, especially targeted but not limited to unsupervised learning. The DeepMI formulation/function is derived intuitively from the expression of Mutual Information ($\mathcal{MI}$) which essentially provides a measure of similarity between two random signals. The computations of DeepMI are entirely based on discrete probability density functions (pdf). The discrete pdf however, doesn't allow backward gradient flow. Hence, a fuzzification strategy to compute fuzzy pdf is also proposed which ensures the end-to-end differentiable optimization. One advantage of DeepMI is that it can be used to train a deep neural network for several tasks which otherwise shall require a different objective function for each task. We choose a few supervised/unsupervised learning tasks and conduct experiments in order to show the importance and competence of DeepMI function over the existing loss functions.
\end{abstract}

\end{justify}

% The regular MI expression, despite being convex, can not be used directly for this purpose due to its unbounded above nature i.e. $\sup{MI}\to\infty$ when two signals match. This nature clearly doesn't depict the stopping criteria for the training process when using MI as the loss function. The DeepMI on the other hand, is minimized when two random signals matches with each other perfectly.

\section{Introduction}
Selection of a suitable loss function is crucial in order to train a neural network for a desired task. For a given neural network architecture \cite{resnet,vgg} and optimization procedure \cite{kingma2014adam}, the profile of the loss function largely governs what is learnt by the neural network and its generalization on unseen data. The above statement well applies to the choice of a similarity metric in the learning process. The deep learning based approaches involve minimizing a similarity metric between the ground truth and the predictions in a way or another. Though, the way it is performed may vary from task to task. For example, the visual perception tasks such as image classification \cite{deng2009imagenet,vgg}, image segmentation \cite{kumar2019semi, pspnet}, it is performed using cross entropy, whereas, for the other tasks such as forecasting \cite{oord2016wavenet}, image generation \cite{gan}, depth estimation \cite{godard2017unsupervised}, it is performed using $\mathcal{L}_2$, $\mathcal{L}_1$ metrics. Generative adversaries \cite{gan} based learning algorithms also largely depends on the choice of similarity metric.
\par
In the area of machine vision, there are certain tasks for which the groundtruth can be not be obtained easily. It is primarily due to a very high cost of measurement devices or unavailability of labelling process. These tasks include depth estimation using images \cite{godard2017unsupervised}, visual / LiDAR odometery \cite{kumar2018real}. From the perspective of the autonomous vehicles and robotics, the above tasks are undeniably important. For this reason, development of unsupervised learning techniques for these tasks have recently gained attention \cite{godard2017unsupervised, zhan2018unsupervised, li2018undeepvo, almalioglu2019ganvo}. The proposed methods in this direction extensively use similarity metric minimization between various information sources. 
\par
From the above discussion, it is quite evident that similarity metric plays an important role in the learning process. In general, $\mathcal{L}_1,\mathcal{L}_2$ are the most preferred choice for this purpose. These losses, despite their popularity, do not provide the desired results in many cases. It is mainly due to the fact that these are pointwise operators and do not account for statistical information while matching. For example, in images, $\mathcal{L}_1/\mathcal{L}_2$ loss penalizes the neural network on a per-pixel basis. The statistical properties are thus left unaccounted. To address this issue, Structural Similarity (SSIM) index \cite{ssim} has recently become popular and is being used as an alternative to $\mathcal{L}_1,\mathcal{L}_2$. The SSIM index is computed over a window instead of a pixel and is based on local statistics. In practice, the aforementioned losses are used in conjunction with each other which increases the number of individual loss functions, leading to an increased complexity in order to tune the loss weights \cite{janai2018unsupervised}. 
\par
Keeping in mind the above observations, in this work, we explore the potential of $\mathcal{MI}$ \cite{kl,mi,entropy} to train deep neural networks for supervised / unsupervised learning of a task. $\mathcal{MI}$ is essentially an information theoretic measure to reason about the statistical independence of two random variables. An interesting property of $\mathcal{MI}$ is that it operates on probability distributions instead of the data directly. Therefore, $\mathcal{MI}$ does not depend on the signal type i.e. images or time-series and proves to be a powerful measure in many areas. For this reason, we consider $\mathcal{MI}$ as a potential alternative measure of similarity. Despite, its diverse applications, the expression of $\mathcal{MI}$ is infeasible to be used directly for the training a neural network (Sec.~\ref{sec:mi}). However, the interesting properties of $\mathcal{MI}$ encourage us to dive deep into the problem and lead us to contribute through this paper as follows:
\begin{itemize}
\item Feasibility of $\mathcal{MI}$ formulation for deep learning tasks.
\item An $\mathcal{MI}$ inspired novel loss function $\mathcal{L}_{LMI}$.
\item DeepMI: a fuzzy logic based framework to train DNNs using $\mathcal{L}_{LMI}$ or $\mathcal{L}_{MI}$.
\end{itemize}
%\par
In the next section, we discuss the related work. In Sec.~\ref{sec:mi}, we brief $\mathcal{MI}$ and in Sec. ~\ref{sec:deepmi}, we discuss the limitations of regular $\mathcal{MI}$ expression and develop $\mathcal{L}_{LMI}$ along with gradient calculation required for back propagation. In Sec.~\ref{sec:exp}, we experimentally verify the importance of DeepMI through a number of unsupervised learning tasks. Finally, Sec.~\ref{sec:conc} provides conclusion about the paper.
\section{Related Work}
Literature on Mutual Information is diverse and vast. Therefore, we limit our discussion only to the most relevant works in this area. Mutual Information \cite{entropy,mi,kl} is a fundamental measure of information theory which provides a sense of independence between random variables. It has been widely used in a variety of applications. The works \cite{viola1997alignment,maes1997multimodality} are typical examples which exploit $\mathcal{MI}$ in order to align medical images. $\mathcal{MI}$ has also been successfully used in speech recognition \cite{bahl1986maximum}, machine vision and robotic applications. \cite{pandey2012toward} is a typical example in the area of autonomous vehicles to register 3D points clouds obtained by LIDARs. Apart from that, $\mathcal{MI}$ has widely been used in independent component analysis \cite{hyvarinen2000independent}, key feature selection \cite{kwak2002input,peng2005feature}. From the above applications of $\mathcal{MI}$ into diverse area, $\mathcal{MI}$ can be thought as a pivotal measure.
\par
The works \cite{viola1997alignment, maes1997multimodality, pandey2012toward} are non-parametric approaches which maximize $\mathcal{MI}$ to achieve the desired purpose. $\mathcal{MI}$ is operated upon the distributions of the raw signals or the features extracted. For example, \cite{viola1997alignment} used image histograms, whereas \cite{pandey2012toward} uses the distributions of 3D points in a voxel. In these techniques the feature extraction is quite important which is handcrafted. In the past decade, deep neural network architectures \cite{resnet, vgg} have been proved to be excellent in learning high quality embeddings / feature from the input data in an entirely unsupervised manner which in turn are used for various tasks \cite{rcnn, fastrcnn, fasterrcnn, maskrcnn,ssd, pspnet,gan}. Therefore, we believe that bringing  deep learning framework in conjunction with $\mathcal{MI}$ can be extremely useful. However, so far, there does not exist any unified standard framework which can be used for this purpose. It is mainly due to the issues related with $\mathcal{MI}$. For example, the distributions required for mutual information are not exact, instead they are only the approximations of the true distribution \cite{darbellay1999estimation}. Also, these approximations are not differentiable, thus making it difficult for $\mathcal{MI}$ to be included in the deep learning methods \cite{belghazi2018mine}. Since, affordable deep learning methods have only recently emerged, the learning process is mostly based on the traditional losses \cite{janai2018unsupervised,godard2017unsupervised, zhan2018unsupervised, li2018undeepvo, almalioglu2019ganvo}. A very recent work \cite{belghazi2018mine} proposes to use $\mathcal{MI}$ with neural networks. However, the work mainly addresses to estimate the distributions using neural networks and does not talk about per-sample $\mathcal{MI}$ which is required for the tasks such as \cite{viola1997alignment, maes1997multimodality, pandey2012toward}.
\par
The works \cite{godard2017unsupervised, zhan2018unsupervised, li2018undeepvo, almalioglu2019ganvo,bian2019unsupervised} in the area of depth estimation and visual odometery using deep neural networks in an unsupervised fashion are typical examples where $\mathcal{MI}$ can be employed. These works only utilize the losses such as $\mathcal{L}_1, \mathcal{L}_2, \mathcal{L}_{SSIM}$. We believe that since $\mathcal{MI}$ has successfully been employed in diverse applications, it is worth developing a well defined and benchmarked $\mathcal{MI}$ based framework for deep learning. Based on the motivation, in this paper, we explore the possibility and feasibility of the above idea of using $\mathcal{MI}$ for robotics applications. Our intention is not to outcast the existing losses, instead to bring in $\mathcal{MI}$ into deep learning, and to build a baseline in the paper to open the doors to new research area in this direction.
\section{Mutual Information ($\mathcal{MI}$)}
\label{sec:mi}
For any two random variables $X$ and $Y$, the measure $\mathcal{MI}$ is defined as.%
\begin{equation}
\mathcal{I}(X;Y) = \mathcal{H}(X) + \mathcal{H}(Y) - \mathcal{H}(X,Y),~~~~~ \mathcal{I}(X;Y) \geq 0
\label{eq_mi}
\end{equation}
\begin{equation}
\begin{aligned}
\mathcal{H}(X) &= -\sum_{x\in X }p_X^x log(p_X^x), \\
\mathcal{H}(Y) &= -\sum_{y\in Y }p_Y^y log(p_Y^y), \\
\mathcal{H}(X,Y) &= -\sum_{x\in X }\sum_{y\in Y }p_{XY}^{xy} log(p_{XY}^{xy})
\end{aligned}
\label{eq_hxhyhxy}
\end{equation}
\par
where $\mathcal{H}(X), \mathcal{H}(Y)$ represent the entropy \cite{entropy} of $X$, entropy of $Y$, whereas $\mathcal{H}(XY)$ represents the joint entropy of $X,Y$ when both variables are co-observed. The symbols $p_X, p_Y$ and $p_{XY}$ represent the marginal of $X$, marginal of $Y$ and joint probability density function (pdf) of $X,Y$ respectively.
\par
Mutual Information is an important term in the information theory as it provides a measure of statistical independence between two random variables based on the distribution. In other words, $\mathcal{MI}$ governs that how well one can explain about a random variable $X$ after observing another random variable $Y$ or vice-versa. The expression of $\mathcal{MI}$ in Eq.~\ref{eq_mi} is defined in terms of entropies. For any random variable $X$, its entropy quantifies uncertainty associated with its occurrence.
\subsection{$\mathcal{MI}$ as a similarity metric}
$\mathcal{MI}$ is a convex function and attains global minima when any two random variables under consideration are independent. Mathematically, ${\mathcal{MI}} \to 0$ when the variables are independent, whereas, ${\mathcal{MI}\to \{\mathcal{H}(X) =\mathcal{H}(Y)\}}$ when both the variables are identical statistically. This property of $\mathcal{MI}$ can readily be employed to quantify similarity between two signals. However, while doing so, the definition of $\mathcal{MI}$ has to be interpreted in a quite different manner.
\par
To better understand, let us consider an example of image matching, provided two images $X$ and $Y$. In order to measure the similarity between the images using $\mathcal{MI}$, the image itself can not be considered as a random variable. Because in that case, $p_X, p_Y$ and $p_{XY}$ shall be meaningless. In other words, per sample $\mathcal{MI}$ is not defined. Hence, instead of an image, its pixel values are considered as a random variable over which the relevant distributions can be defined. The pixel values may refer to intensity, color, gradients etc. In order to compute the similarity score, first the marginals and joint pdfs over the selected variable has to be computed, and the similarity can be obtained by using the Eq.~\ref{eq_mi}. While doing so, Eq.~\ref{eq_hxhyhxy} needs to be rewritten as given below.
\begin{equation}
\begin{aligned}
\mathcal{H}(X) &= -\sum_{i=1}^{N}p_X^i log(p_X^i), \\
\mathcal{H}(Y) &= -\sum_{i=1}^{N}p_Y^i log(p_Y^i),\\
\mathcal{H}(X,Y) &= -\sum_{i=1}^{N}\sum_{j=1}^{N}p_{XY}^{ij} log(p_{XY}^{ij})
\end{aligned}
\label{eq_hxhyhxyps}
\end{equation}
Where $N$ is the number of bins in the pdf.
\par
As an another example, we can consider matching of two time series signals by using $\mathcal{MI}$. Following the above discussion, the two signal instances under consideration can not be considered as random variables, instead their instantaneous values are considered as random variable.  It must be noticed that the choice of random variable depends on the application.
\section{DeepMI Framework}
\label{sec:deepmi}
To understand the concept of DeepMI, consider the task of image reconstruction using autoencoders. In order to minimize the gap between an input image and the reconstructed image, the $\mathcal{MI}$ has to be maximized. The regular $\mathcal{MI}$ expression however, can not be used directly for this purpose. It is primarily because $\mathcal{MI}$ attains global minima when both random variables are dissimilar and our optimal point which is $\sup~{\mathcal{MI}}$, is not well defined as $\mathcal{MI}$ is unbounded above. Although, various normalized versions of $\mathcal{MI}$ have also been proposed \cite{nmi} in the literature, the previously discussed issues still remain intact. Hence, normalized $\mathcal{MI}$ ($\mathcal{NMI}$) also can not serve our purpose.
%\begin{equation}
%\mathcal{NMI} = \frac{\mathcal{MI}}{max(\mathcal{H}(X),\mathcal{H}(Y))},~~~~~\mathcal{NMI} = \frac{2 * \mathcal{MI}}{\mathcal{H}(X) + \mathcal{H}(Y)}
%\end{equation}
%
\par
The above challenges encourage us to develop linearized mutual information $\mathcal{LMI}$ which attains a global minima when the two images are exactly the same. In order to achieve this, we turn towards the the working of the $\mathcal{MI}$ and make a following important insight. 
%The above expressions are derived by leveraging the loose upperbounds of $\mathcal{MI}$ which are given by $\mathcal{I}(X;Y) \leq \mathcal{H}(X)$ and $\mathcal{I}(X;Y) \leq \mathcal{H}(Y)$. 
%
\subsection{A key insight to MI}
Consider two images $X$ and $Y$, with $p_X \in \mathbb{R}^{N}, p_Y \in \mathbb{R}^{N}$ and $p_{XY} \in \mathbb{R}^{N \times N}$ as their marginals and joint pdfs respectively. The dimensions of $p_X, p_Y$ is $N\times 1$, whereas it is $N\times N$ for $p_{XY}$. From, Eq.~\ref{eq_mi},~\ref{eq_hxhyhxy}, we can immediately say that $\mathcal{MI}\to 0$ when two signals are dissimilar while ${\mathcal{MI}\to \{\mathcal{H}(X) =\mathcal{H}(Y)\}}$ when the signals are exactly the same.  Hence, for the images $X$ and $Y$ to be identical, the necessary but not sufficient condition is that $p_X$ and $p_Y$ should be same. While, in order to guarantee, the following has to be satisfied.
\begin{equation}
p_X^i = p_Y^i = p_{XY}^{ii},~~~ p_{XY_{\vert i \neq j}}^{ij} = 0,~~~~~ i,j=1,2,...,N
\end{equation}
In other words, when $X \equiv Y$, the off-diagonal elements of $p_{XY}$ are zero while all the diagonal elements are non-zero (depending on a distribution) and equals to $p_{X}$ and $p_Y$ simultaneously. The above insight leads us to derive an expression for the $\mathcal{LMI}$ function to train deep networks.
\subsection{$\mathcal{LMI}$ Derivation}
We know that for any probability density function Eq.~\ref{eq_pdfsum} and \ref{eq_pdfsum1} holds. These equations represent a $1$D and a $2$D probability density function respectively.

%\footnotesize
\begin{align}
\footnotesize
\sum_{i=1}^{N}p^i_X = \sum_{i=1}^{N}p^i_Y =1, \label{eq_pdfsum}
\\
\sum_{i=1}^{N}\sum_{j=1}^{N}p^{ij}_{XY}=1 
\label{eq_pdfsum1}
\end{align}
%\normalsize
%
Rewriting Eq.~\ref{eq_pdfsum1} as combination of its diagonal $(i=j)$ and off-diagonal elements $(i\neq j)$, we get
\begin{align}
& \sum_{i=1}^{N}\sum_{\substack{j=1\\i \neq j}}^{N}p_{XY}^{ij} + \sum_{i=1}^{N}p_{XY}^{ii} =1
\label{eq:diagoffdiag} 
\\
& \sum_{i=1}^{N}\sum_{\substack{j=1\\i \neq j}}^{N}p_{XY}^{ij} + \sum_{i=1}^{N}p_{XY}^{ii} ~+ \sum_{i=1}^{N}p_{XY}^{ii} - \sum_{i=1}^{N}p_{XY}^{ii} =1 
\end{align}
\begin{equation}
\begin{aligned}
&\sum_{i=1}^{N}\sum_{\substack{j=1\\i \neq j}}^{N}p_{XY}^{ij} + \sum_{i=1}^{N}p_{XY}^{ii} ~+  \sum_{i=1}^{N}p_{XY}^{ii} - \sum_{i=1}^{N}p_{XY}^{ii} + \\
& \sum_{i=1}^{N}p_{X}^{i} -
\sum_{i=1}^{N}p_{X}^{i} +
\sum_{i=1}^{N}p_{Y}^{i} -
\sum_{i=1}^{N}p_{Y}^{i} = 1 
\end{aligned}
\end{equation}
\begin{equation}
\begin{aligned}
& \sum_{i=1}^{N}\sum_{\substack{j=1\\i \neq j}}^{N}p_{XY}^{ij} + \sum_{i=1}^{N}\vert p_{XY}^{ii} -p_{X}^{i}\vert ~+ \\
& \sum_{i=1}^{N}\vert p_{XY}^{ii} -p_{Y}^{i}\vert -
\sum_{i=1}^{N}p_{XY}^{ii} +
1+
1  \geq 1 
\end{aligned}
\end{equation}
\begin{equation}
\begin{aligned}
& \sum_{i=1}^{N}\sum_{\substack{j=1\\i \neq j}}^{N}p_{XY}^{ij} + \sum_{i=1}^{N}\vert p_{XY}^{ii} -p_{X}^{i}\vert ~+ \\
& \sum_{i=1}^{N}\vert p_{XY}^{ii} -p_{Y}^{i}\vert -
\sum_{i=1}^{N}p_{XY}^{ii} + 1  \geq 0 
\label{eq:deepmiderivation}
\end{aligned}
\end{equation}
Now, reffering to Eq.~\ref{eq:diagoffdiag}, we can write
\begin{equation}
 \sum_{i=1}^{N}p_{XY}^{ii}  \leq 1 ~~~\Rightarrow~~~
1- \sum_{i=1}^{N}p_{XY}^{ii}  \geq 0
\label{eq:pxybound}
\end{equation}
Using the above into Eq.~\ref{eq:deepmiderivation}, we get
\begin{equation}
\begin{aligned}
& \mathcal{L}_{LMI} = \frac{1}{3} \Big( \sum_{i=1}^{N}\sum_{\substack{j=1\\i \neq j}}^{N}p_{XY}^{ij} + \\ & \sum_{i=1}^{N}\vert p_{XY}^{ii} -p_{X}^{i}\vert +
\sum_{i=1}^{N}\vert p_{XY}^{ii} -p_{Y}^{i}\vert \Big)  & \geq 0 
\end{aligned}
\label{eq:deepmi}
\end{equation}
Where the factor $\nicefrac{1}{3}$ is included to ensure $\mathcal{LMI} \leq 1$. It is obtained by replacing each of the three terms to their maximum. The equality of the equation to $0$ will hold \textit{iff} $p_X^i = p_Y^i = p_{XY}^{ii},~~p_{XY_{\vert i \neq j}}^{ij}=0~~\forall~i,j \in 1,2,..,N$. i.e. two images match perfectly. Hence, the L.H.S. of the equation Eq.~\ref{eq:deepmi} is treated as the objective function which we call the $\mathcal{LMI}$ function. The ``$\mathcal{L}$'' stands for ``linearized'' which arises because the $\mathcal{LMI}$ formulation is linear in the elements of the pdfs, whereas the ``$\mathcal{MI}$'' term arises because at the equality, the regular $\mathcal{MI}$ expression will also be maximized. The $\mathcal{LMI}$ formulation is quite interesting because it is essentially a combination of three different losses weighted equally. The $\mathcal{L}_{LMI}$ formulation is quite intuitive and the gradients are straightforward to compute.
\subsection{Fuzzy Probability Density Function}
The $\mathcal{LMI}$ function utilizes the pdfs $p_X,p_Y$ and $p_{XY}$ which are discrete in nature. These are typically obtained by computing an $N$ bin histogram followed by a normalization step with $||.||_1=1$. As per the standard procedure to compute a regular histogram, first, a bin-id corresponding to an observation of the random variable is computed and later, the count of the respective bin is incremented by unity. The computation of the bin-id is carried out by a $ceil$ or $floor$ operation which is not differentiable. While performing the rounding step, the actual contribution of the observation is lost. Thus, both the incremental and rounding procedure prevent the gradient flow which is needed during the training process.
\par
To better understand the above, let $h_X$ be an $N$ bin histogram of the random variate $X$ and $x\in X$ be an observation. In the case of a regular histogram, the bin-id $x_b$ corresponding to an observation $x$ is computed as follows:
\begin{equation}
\begin{aligned}
x_b = \frac{\hat{x}}{bin\_res},~~~& ~~~  \hat{x} = \frac{x-min_X}{max_X-min_X}, \\ bin\_res &= \frac{max_X-min_X}{N} 
\end{aligned}
\label{eq:bin}
\end{equation}
where $max_X, min_X$ are the maximum and minimum values which the variable $X$ can attain at any instant. Typically for $8$-bit images, $[min_X,max_X]=[0,255]$. From the Eq.~\ref{eq:bin}, it can be noticed that the value of $x_b$ is not necessarily an integer. In this case, $x_b$ is rounded to the nearest integer by using $ceil\equiv \lceil x_b \rceil$ or $floor \equiv \lfloor x_b \rfloor$, depending upon one's convention. Now, the count of $x_b$ is incremented by one. Therefore, it becomes evident that the rounding procedure and the unit incremental procedure do not allow the gradient computation w.r.t. the observations. To cope up with this, we employ fuzzification strategy in order to ensure valid gradients during back-propagation.
\subsubsection{Fuzzification of $p_X$, $p_Y$}
In order to fuzzify $h_X$, instead of one as in Eq.~\ref{eq:bin}, we compute two bins corresponding to $x\in X$ i.e. $x_0 = \lfloor x_b \rfloor$ and $x_1 = \lceil x_b \rceil$. We define a membership function for each of the two bins as follows.
\begin{equation}
m_{x_0} = 1-(x_b-x_0),~~~m_{x_1} = (x_b-x_0) 
\label{eq:mempxpy}
\end{equation}
where $m_{x_0}, m_{x_1}$ are the membership functions of $x_0$ and $x_1$. Essentially, $m_{x_0} + m_{x_1} = 1$. While performing the unit incremental step, the count of the bins $x_0$ and $x_1$ are incremented by $m_{x_0}$ and $m_{x_1}$ respectively instead of increasing by one. With the help of fuzzification, it can be inferred that now the gradients of $x_b$ w.r.t. $m_{x_0}$ and $m_{x_1}$ are fully defined (Sec.~\ref{sec:derivatives}). The above steps are followed in order to compute $p_X$ and $p_Y$, while the normalization step being performed at the end. As a matter of convention, the memberships corresponding to $y_b$ for a $y\in Y$ are denoted by $m_{y_0}$ and $m_{y_1}$. 

\subsubsection{Fuzzification of $p_{XY}$}
The fuzzification of a joint pdf $p_{XY}$ is simply an extension of previous steps. For a regular $2$D histogram, the unit incremental procedure is applied to the bin location defined by the coordinate $(x_b,y_b)$ (Eq.~\ref{eq:bin}) which in this case as well, need not to be exactly an integral value. Therefore, to ensure valid gradient flow in this case, four memberships are defined which corresponds to the four coordinates top-left, top-right, bottom-left, and bottom-right w.r.t $(x_b,y_b)$. Mathematically, these four coordinates are given by $(x_0,y_0)$, $(x_1,y_0)$, $(x_0,y_1)$, $(x_1,y_1)$, and their respective memberships can be written as:
\begin{equation}
\begin{aligned}
&m_{x_0 y_0} = m_{x_0}m_{y_0},~~~
m_{x_1 y_0} = m_{x_1}m_{y_0},\\
&m_{x_0 y_1} = m_{x_0}m_{y_1},~~~
m_{x_1 y_1} = m_{x_1}m_{y_1}
\end{aligned}
\label{eq:mempxy}
\end{equation}
While performing the unit incremental step, the count of the four bins mentioned above is incremented by their respective membership value.
\subsection{Back-propagation through DeepMI framework}
\label{sec:derivatives}
From the Eq.~\ref{eq:mempxy}, it can be inferred that, gradient of $x$ w.r.t. $\mathcal{L}_{LMI}$ depends on $p_X^{x_0},p_X^{x_0},p_{XY}^{x_0y_0},p_{XY}^{x_0y_1},
p_{XY}^{x_1y_0},p_{XY}^{x_1y_1}$. Therefore, we can write:
\begin{equation}
\frac{\partial \mathcal{L}}{\partial x} = \sum_{i=0}^{1} \frac{\partial \mathcal{L}}{\partial p_{X}^{x_i}}~ \frac{\partial p_{X}^{x_i}}{\partial x}  + \sum_{i=0}^{1}\sum_{j=0}^{1}\frac{\partial \mathcal{L}}{\partial p_{XY}^{x_iy_i}}~ \frac{\partial p_{XY}^{x_iy_i}}{\partial x}
\end{equation}
Using chain rule:
\begin{equation}
\begin{aligned}
 \frac{\partial p_{X}^{x_i}}{\partial x} =  \frac{\partial p_{X}^{x_i}}{\partial \hat{p}_{X}^{x_i}} \times \frac{\partial \hat{p}_{X}^{x_i}}{\partial {m_{x_i}}} \times \frac{\partial m_{x_i}}{\partial x_b} \times \\ \frac{\partial x_b}{\partial \hat{x}} \times \frac{\partial \hat{x}}{\partial x},~~p_{X}^{x_i} = \frac{\hat{p}_{X}^{x_i}}{\sum_{i=1}^{N}\hat{p}_{X}^{x_i}}
\end{aligned}
\end{equation}
and similarly,
\begin{equation}
\begin{aligned}
 \frac{\partial p_{XY}^{x_iy_j}}{\partial x} =  \frac{\partial p_{XY}^{x_iy_j}}{\partial \hat{p}_{XY}^{x_iy_j}} & \times \frac{\partial \hat{p}_{XY}^{x_iy_j}}{\partial m_{x_iy_j}} \times \frac{\partial m_{x_iy_j}}{\partial m_{x_i}} \times \frac{\partial m_{x_i}}{\partial x_b} \\ & \times \frac{\partial x_b}{\partial \hat{x}} \times \frac{\partial \hat{x}}{\partial x}, ~~~ p_{XY}^{x_iy_j} = \frac{\hat{p}_{XY}^{x_iy_j}}{\sum_{i=1}^{N}\hat{p}_{XY}^{x_iy_j}}
\end{aligned}
\end{equation}
Each of the above partial derivatives can be easily computed using the Eq.~\ref{eq:deepmi}~-~\ref{eq:mempxy} via the chain rule. It must be noticed that in the gradient calculations, $x_0,x_1$ does not occur which would have been there in case of regular histograms, leading to undefined derivatives. Similarly, gradients can be also be computed for an observation $y$.
\subsection{$\mathcal{L}_{LMI}$ Implementation}
The formulation of $\mathcal{L}_{LMI}$ seems quite intuitive. There is however a consideration which must be accounted while using it for signal matching, especially in the scenarios where one of the signal is the groundtruth and the another is the estimated version of the first by a neural network.
\par
To understand this, consider two images $X$ and $Y$, where $X$ is the image to be reconstructed and $Y$ is the reconstruction or simply the output of a neural network. The marginals $p_X,p_Y$ and the joint $p_{XY}$ of $X$ and $Y$ are computed using the fuzzification procedure as described previously. As we know that these distributions are the approximated version of the underlying distribution, therefore, undersampling or oversampling of the underlying distribution is possible. By nature the distributions $p_X$, $p_Y$ and $p_{XY}$ do not have a well defined mathematical expressions in such scenarios. Moreover, both the $p_X$ and $p_Y$ can be obtained using $p_{XY}$, therfore, while backpropagation, the gradients only w.r.t $p_{XY}^{ij}$ are backpropagated. Mathematically, gradients w.r.t. $p_Y$ should also be backpropagated because calculations of $p_Y$ is dependent on $Y$. However doing so, disturbs the training process and affects the neural network performance. After our experimental study, we mark this observation as an outcome of the distributions approximation procedure. 
\subsection{Hyperparameters}
\label{sec:hyper}
The DeepMI framework has $N$ as the only hyperparameter. The number of bins $N$ mainly depicts that how precisely the $\mathcal{L}_{LMI}$ should penalize the network while matching. For example, consider an $8$-bit image with dynamic range $[0,255]$. Now, we set $N=255$, the network will be penalized very strongly while matching.  While if the $N$ is set to a small number, the network will be forced to focus only on the important details. This property can be quite useful in cases where two images from different cameras need to be matched while both the images have different brightness levels.
% Moreover, we also define a window $K_1\times K_2$, over which the various pdfs are computed. For an image signal, the window size may range from smaller windows with a given stride to the whole image. The same applies for a time-series signal, with $K_2=1$. This gives us the flexibility to match the information locally or globally. To understand better, we study the effect of each hyperparameter in Sec.~\ref{sec:exphparams}.
%
%
\section{Experiments}
\label{sec:exp}
In this section, we benchmark the effectiveness and the applicability of the DeepMI framework. Due to the unavailability of a standard evaluation procedure, we define three tasks on which a deep neural network is trained in an unsupervised manner. The three tasks vary in their difficulty levels from baseline, moderate, to extremely difficult from the learning perspective. As an experimental study, the training of each task is performed using different loss functions $\mathcal{L}_1, \mathcal{L}_2, \mathcal{L}_{SSIM}$ along with $\mathcal{L}_{LMI}$. For $\mathcal{L}_{SSIM}$, we use a $3 \times 3$ block filter following \cite{godard2017unsupervised}. All the performance metrics are provided in Table-\ref{tab:unified} and Table-\ref{tab:effectN} for quantitative analysis with the best scores highlighted in the \textcolor{blue}{blue}. For training, we use \texttt{base\_lr} = $0.001$, \texttt{lr\_policy}=$poly$, \texttt{ADAM} optimizer with $\beta_1=0.9$ and $\beta_2=0.99$ unless otherwise stated. The encoder-decoders \cite{unet} have a filter size of $3\times3$ and the number of filters equals to $16,32,64,128,256$ for the five stages for each task. The whole framework has been implemented as a layer-wise architecture in C++ and the codes will be available at the link provided in the beginning.
%
%\subsection{Autoencoders}
%Autoencodings are the one of machine learning techniques for feature leaning in an unsupervised manner. In this experiment we, perform autoencoding of  using neural netoworks. We study the effects of using various loss funcions and compare them with DeepMI. The reconstruction quality will be used to asses the  performance of the objective function. During all the experiments, network architecture and other optimzation hyperparameters are is fixed. 
%After experimental observation, we find all the looses perfrms equally for the task of autoencoding.
%
\subsection{Unsupervised Bar Alignment (\texttt{Exp$_1$})}
This experiment consists of two binary images where the second image is a spatially transformed version of the first i.e. a $2$D rigid body transform is defined between the two images. Both the images consist of a black background and a white rectangular bar. The bar sized of $50\times 125$ in an image ($192\times
640$) has two degrees-of-freedom (DoF): $tx$, and $\theta$ which correspond to horizontal motion and in-plane rotation. The goal of the experiment is to learn the $2$DoF parameters in an unsupervised manner. For the training purpose, we generate a dataset of $1500$ images with $1000+500$ train-test split.  The dataset is generated by transforming the bar in the image by randomly generating $tx \in [-100,100]$ pixels and $\theta \in [-40,40]$ degrees. The training of this experiment is performed using \texttt{SGD} with Nesterov momentum $=0.95$ for $5$ epochs.
\par
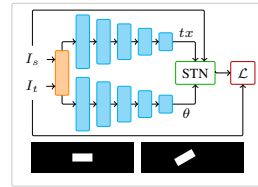
\begin{figure}[t]
\centering

\colorlet{conv}{white!60!cyan}
\colorlet{concat}{white!60!orange}
\begin{tikzpicture}
\node (outer) [draw=white!80!black,rounded corners=0.25mm,scale = 0.75]{
\tikz{
\node (s1) [draw=cyan, fill=conv, minimum width=1.5ex,rounded corners=0.25mm, minimum height = 6ex, xshift = -4 ex, yshift = 0 ex] {};
\node (s2) [draw=cyan, fill=conv,right of = s1,rounded corners=0.25mm,minimum width=1.5ex, minimum height = 5ex, xshift = -4ex] {};
\node (s3) [draw=cyan, fill=conv,right of = s2,rounded corners=0.25mm,minimum width=1.5ex, minimum height = 4ex, xshift = -4ex] {};
\node (s4) [draw=cyan, fill=conv,right of = s3,rounded corners=0.25mm,minimum width=1.5ex, minimum height = 3ex, xshift = -4ex] {};
\node (s5) [draw=cyan, fill=conv,right of = s4,rounded corners=0.25mm,minimum width=1.5ex, minimum height = 2ex, xshift = -4ex] {};
\draw [->,very thin] (s1) -- (s2);
\draw [->,very thin] (s2) -- (s3);
\draw [->,very thin] (s3) -- (s4);
\draw [->,very thin] (s4) -- (s5);
\node (s12) [draw=cyan, fill=conv, minimum width=1.5ex,rounded corners=0.25mm, minimum height = 6ex, xshift = -4 ex, yshift = -7 ex] {};
\node (s22) [draw=cyan, fill=conv,right of = s1,rounded corners=0.25mm,minimum width=1.5ex, minimum height = 5ex, xshift = -4ex, yshift = -7 ex] {};
\node (s32) [draw=cyan, fill=conv,right of = s2,rounded corners=0.25mm,minimum width=1.5ex, minimum height = 4ex, xshift = -4ex, yshift = -7 ex] {};
\node (s42) [draw=cyan, fill=conv,right of = s3,rounded corners=0.25mm,minimum width=1.5ex, minimum height = 3ex, xshift = -4ex, yshift = -7 ex] {};
\node (s52) [draw=cyan, fill=conv,right of = s4,rounded corners=0.25mm,minimum width=1.5ex, minimum height = 2ex, xshift = -4ex, yshift = -7 ex] {};
\draw [->,very thin] (s12) -- (s22);
\draw [->,very thin] (s22) -- (s32);
\draw [->,very thin] (s32) -- (s42);
\draw [->,very thin] (s42) -- (s52);
\node (stn) [draw=green!70!black,right of = s5,minimum width=2.5ex, minimum height = 2ex, xshift = -3 ex, yshift=-3.5ex]{\scriptsize STN};
\draw [->,very thin] (s5) -| (stn) node[xshift=-1ex, yshift=4.5ex]{\scriptsize $tx$};
\draw [->,very thin] (s52) -| (stn) node[xshift=-1ex, yshift=-4.5ex]{\scriptsize $\theta$};
\node (c1) [draw=orange,fill=concat,left of=s1, minimum width=1.5ex,rounded corners=0.25mm, minimum height = 5ex, xshift = 4 ex, yshift = -3.5 ex] {};
\node (i1) [draw=none,left of=c1,minimum width=1.5ex,rounded corners=0.25mm, minimum height = 6ex, xshift = 3 ex, yshift = 1.5 ex] {\scriptsize $I_s$};
\node (i2) [draw=none,left of=c1,minimum width=1.5ex,rounded corners=0.25mm, minimum height = 6ex, xshift = 3 ex, yshift = -1.5 ex] {\scriptsize $I_t$};
\draw [->,very thin] (i1) -- ($(c1.west) + (0ex,1.5ex)$);
\draw [->,very thin] (i2) -- ($(c1.west) - (0ex,1.5ex)$);
\draw [->,very thin] (c1) |- (s1);
\draw [->,very thin] (c1) |- (s12);
\draw [->,very thin] ($(i1.north) - (0ex,1.5ex)$) -- ($(i1.north) + (0ex,2.5ex)$) -| ($(stn.north) + (1ex,0ex)$);
\node (loss) [draw=red!70!black,right of = stn,minimum width=2.5ex, minimum height = 2ex, xshift = -1 ex, yshift=0ex]{\scriptsize $\mathcal{L}$};
\draw [->,very thin] (stn) |- (loss);
\draw [->,very thin] ($(i2.south) + (0ex,1.5ex)$) -- ($(i2.south) - (0ex,2.5ex)$) -| (loss);
\node (img1) [draw=none,minimum width=1ex, minimum height = 2ex, xshift = -4 ex, yshift=-13ex]{\includegraphics[scale=0.08]{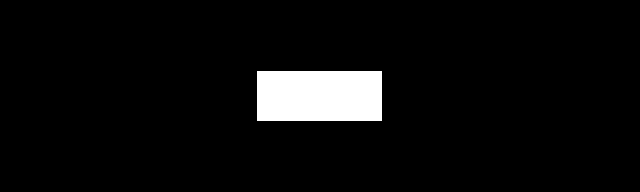}};
%\\
%
\node (img2) [draw=none, right of = img1, minimum width=1ex, minimum height = 2ex, xshift = 6 ex, yshift=0ex]{\includegraphics[scale=0.08]{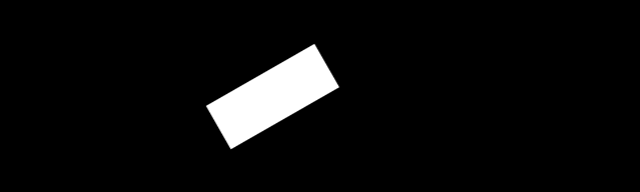}};
}};
\end{tikzpicture}
\caption{Unsupervised learning framework for the \texttt{Exp$_1$}. The blue box is a convolutional block, orange box is a concatenation block. STN~\cite{jaderberg2015spatial}, and $\mathcal{L}$ is a loss function.}
\label{fig:exp1}
\end{figure}
\begin{figure}[t]
\centering
\begin{tikzpicture}
\FPeval{xoffset}{8.9}
\FPeval{yoffset}{3.1}
\FPeval{imscale}{0.059}

\foreach \i in {0,1,...,4}
{
\node (bai\i) [draw=none,rounded corners=0.6mm, xshift=0 * \xoffset ex, yshift=-(\i-1)*\yoffset ex]{\includegraphics[scale= \imscale]{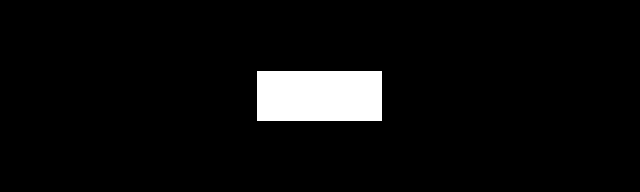}};
}
\foreach \i in {0,1,...,4}
{
\node (bat\i) [draw=none,rounded corners=0.6mm, xshift=1 * \xoffset ex, yshift=-(\i-1)*\yoffset ex]{\includegraphics[scale= \imscale]{baralign_t_l1_\i.png}};
}
\foreach \i in {0,1,...,4}
{
\node (bat\i) [draw=none,rounded corners=0.6mm, xshift=2 * \xoffset ex, yshift=-(\i-1)*\yoffset ex]{\includegraphics[scale= \imscale]{baralign_a_l1_\i.png}};
}
\foreach \i in {0,1,...,4}
{
\node (bat\i) [draw=none,rounded corners=0.6mm, xshift=3 * \xoffset ex, yshift=-(\i-1)*\yoffset ex]{\includegraphics[scale= \imscale]{baralign_a_l2_\i.png}};
}
\foreach \i in {0,1,...,4}
{
\node (bat\i) [draw=none,rounded corners=0.6mm, xshift=4 * \xoffset ex, yshift=-(\i-1)*\yoffset ex]{\includegraphics[scale= \imscale]{baralign_a_ssim_\i.png}};
}
\foreach \i in {0,1,...,4}
{
\node (bat\i) [draw=none,rounded corners=0.6mm, xshift=5 * \xoffset ex, yshift=-(\i-1)*\yoffset ex]{\includegraphics[scale= \imscale]{baralign_a_mi_\i.png}};
}
\FPeval{i}{5}
\node (bat\i) [draw=none,rounded corners=0.6mm, xshift=0 * \xoffset ex, yshift=-(\i-1)*\yoffset ex]{Input};
\FPeval{i}{5}
\node (bat\i) [draw=none,rounded corners=0.6mm, xshift=1 * \xoffset ex, yshift=-(\i-1)*\yoffset ex]{Target};
\FPeval{i}{5}
\node (bat\i) [draw=none,rounded corners=0.6mm, xshift=2 * \xoffset ex, yshift=-(\i-1)*\yoffset ex]{$\mathcal{L}_1$};
\FPeval{i}{5}
\node (bat\i) [draw=none,rounded corners=0.6mm, xshift=3 * \xoffset ex, yshift=-(\i-1)*\yoffset ex]{$\mathcal{L}_2$};
\FPeval{i}{5}
\node (bat\i) [draw=none,rounded corners=0.6mm, xshift=4 * \xoffset ex, yshift=-(\i-1)*\yoffset ex]{$\mathcal{L}_{SSIM}$};
\FPeval{i}{5}
\node (bat\i) [draw=none,rounded corners=0.6mm, xshift=5 * \xoffset ex, yshift=-(\i-1)*\yoffset ex]{$\mathcal{L}_{LMI}$};
\end{tikzpicture}
\caption{Qualitative results of \texttt{Exp$_1$}}
\label{exp1}
\end{figure}

The unsupervised training pipeline for the task is depicted in Fig.~\ref{fig:exp1}. While training, the neural network takes two images, source ($I_s$) and target ($I_t$) as input and predicts $tx, \theta$. The image $I_s$ is then warped ($\hat{I}_s$) using fully differentiable spatial-transformer-networks (STN) \cite{jaderberg2015spatial} and $\hat{I}_s$ is obtained. Now, the neural network is penalized using back-propagation to force $\hat{I}_s \to I_t$. In the testing phase, the neural network predicts $tx, \theta$ on the test data and Mean-Absolute-Error (MAE) is reported between the predictions and the groundtruth $tx, \theta$, already stored during the data generation process. From the Table-\ref{tab:unified} under the column \texttt{Exp$_1$}, it can be noticed that the network trained using $\mathcal{L}_{LMI}$ formulation exhibits better performance as compared to the rest of the loss functions. Fig.~\ref{exp1} shows a few qualitative results of this experiment. In this task, almost all of the losses perform equally well. It can be verified visually as well as from the quantitative results provided in the Table~\ref{tab:unified}.

\begin{figure*}[t]
\centering
\begin{tikzpicture}
\FPeval{xoffset}{15.9}
\FPeval{yoffset}{5.1}
\FPeval{imscale}{0.109}

\foreach \i in {0,1,...,4}
{
\node (bai\i) [draw=none,rounded corners=0.6mm, xshift=0 * \xoffset ex, yshift=-(\i-1)*\yoffset ex]{\includegraphics[scale= \imscale]{mp_i_l1_\i.png}};
}
\foreach \i in {0,1,...,4}
{
\node (bat\i) [draw=none,rounded corners=0.6mm, xshift=1 * \xoffset ex, yshift=-(\i-1)*\yoffset ex]{\includegraphics[scale= \imscale]{mp_t_l1_\i.png}};
}
\foreach \i in {0,1,...,4}
{
\node (bat\i) [draw=none,rounded corners=0.6mm, xshift=2 * \xoffset ex, yshift=-(\i-1)*\yoffset ex]{\includegraphics[scale= \imscale]{mp_m_l1_\i.png}};
}
\foreach \i in {0,1,...,4}
{
\node (bat\i) [draw=none,rounded corners=0.6mm, xshift=3 * \xoffset ex, yshift=-(\i-1)*\yoffset ex]{\includegraphics[scale= \imscale]{mp_p_l1_\i.png}};
}
\foreach \i in {0,1,...,4}
{
\node (bat\i) [draw=none,rounded corners=0.6mm, xshift=4 * \xoffset ex, yshift=-(\i-1)*\yoffset ex]{\includegraphics[scale= \imscale]{mp_p_l2_\i.png}};
}
\foreach \i in {0,1,...,4}
{
\node (bat\i) [draw=none,rounded corners=0.6mm, xshift=5 * \xoffset ex, yshift=-(\i-1)*\yoffset ex]{\includegraphics[scale= \imscale]{mp_p_ssim_\i.png}};
}
\foreach \i in {0,1,...,4}
{
\node (bat\i) [draw=none,rounded corners=0.6mm, xshift=6 * \xoffset ex, yshift=-(\i-1)*\yoffset ex]{\includegraphics[scale= \imscale]{mp_p_mi_\i.png}};
}
\FPeval{i}{5}
\node (bat\i) [draw=none,rounded corners=0.6mm, xshift=0 * \xoffset ex, yshift=-(\i-1)*\yoffset ex]{Input};
\FPeval{i}{5}
\node (bat\i) [draw=none,rounded corners=0.6mm, xshift=1 * \xoffset ex, yshift=-(\i-1)*\yoffset ex]{Target};
\FPeval{i}{5}
\node (bat\i) [draw=none,rounded corners=0.6mm, xshift=2 * \xoffset ex, yshift=-(\i-1)*\yoffset ex]{GT Mask};
\FPeval{i}{5}
\node (bat\i) [draw=none,rounded corners=0.6mm, xshift=3 * \xoffset ex, yshift=-(\i-1)*\yoffset ex]{$\mathcal{L}_1$};
\FPeval{i}{5}
\node (bat\i) [draw=none,rounded corners=0.6mm, xshift=4 * \xoffset ex, yshift=-(\i-1)*\yoffset ex]{$\mathcal{L}_2$};
\FPeval{i}{5}
\node (bat\i) [draw=none,rounded corners=0.6mm, xshift=5 * \xoffset ex, yshift=-(\i-1)*\yoffset ex]{$\mathcal{L}_{SSIM}$};
\FPeval{i}{5}
\node (bat\i) [draw=none,rounded corners=0.6mm, xshift=6 * \xoffset ex, yshift=-(\i-1)*\yoffset ex]{$\mathcal{L}_{LMI}$};
\end{tikzpicture}
\caption{Qualitative results of \texttt{Exp$_2$}}

\label{exp2}
\end{figure*}
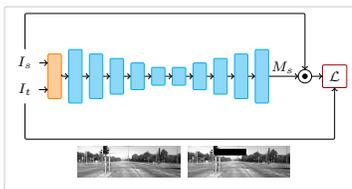
\begin{figure}[t]
\centering

\colorlet{conv}{white!60!cyan}
\colorlet{concat}{white!60!orange}
\begin{tikzpicture}
\node (outer) [draw=white!80!black,rounded corners=0.25mm,scale = 0.75]{
\tikz{
\node (s1) [draw=cyan, fill=conv, minimum width=1.5ex,rounded corners=0.25mm, minimum height = 6ex, xshift = -4 ex, yshift = 0 ex] {};
\node (s2) [draw=cyan, fill=conv,right of = s1,rounded corners=0.25mm,minimum width=1.5ex, minimum height = 5ex, xshift = -4ex] {};
\node (s3) [draw=cyan, fill=conv,right of = s2,rounded corners=0.25mm,minimum width=1.5ex, minimum height = 4ex, xshift = -4ex] {};
\node (s4) [draw=cyan, fill=conv,right of = s3,rounded corners=0.25mm,minimum width=1.5ex, minimum height = 3ex, xshift = -4ex] {};
\node (s5) [draw=cyan, fill=conv,right of = s4,rounded corners=0.25mm,minimum width=1.5ex, minimum height = 2ex, xshift = -4ex] {};
\node (s5d) [draw=cyan, fill=conv,right of = s5,rounded corners=0.25mm,minimum width=1.5ex, minimum height = 2ex, xshift = -4ex] {};
\node (s4d) [draw=cyan, fill=conv,right of = s5d,rounded corners=0.25mm,minimum width=1.5ex, minimum height = 3ex, xshift = -4ex] {};
\node (s3d) [draw=cyan, fill=conv,right of = s4d,rounded corners=0.25mm,minimum width=1.5ex, minimum height = 4ex, xshift = -4ex] {};
\node (s2d) [draw=cyan, fill=conv,right of = s3d,rounded corners=0.25mm,minimum width=1.5ex, minimum height = 5ex, xshift = -4ex] {};
\node (s1d) [draw=cyan, fill=conv,right of=s2d, minimum width=1.5ex,rounded corners=0.25mm, minimum height = 6ex, xshift = -4 ex, yshift = 0 ex] {};
\draw [->,very thin] (s1) -- (s2);
\draw [->,very thin] (s2) -- (s3);
\draw [->,very thin] (s3) -- (s4);
\draw [->,very thin] (s4) -- (s5);
\draw [->,very thin] (s5) -- (s5d);
\draw [->,very thin] (s5d) -- (s4d);
\draw [->,very thin] (s4d) -- (s3d);
\draw [->,very thin] (s3d) -- (s2d);
\draw [->,very thin] (s2d) -- (s1d);
\node (dot) [draw=black,right of = s1d,minimum width=2.5ex, minimum height = 2ex, xshift = -1.5 ex, yshift=0ex,circle,scale=0.6]{};
\node (dot1) [draw=black,fill=black,right of = s1d,minimum width=2.5ex, minimum height = 2ex, xshift = -1.5 ex, yshift=0ex,circle,scale=0.2]{};
\draw [->,very thin] (s1d) -- (dot) node[xshift=-2.5ex, yshift=1ex]{\scriptsize $M_s$};
\node (c1) [draw=orange,fill=concat,left of=s1, minimum width=1.5ex,rounded corners=0.25mm, minimum height = 5ex, xshift = 4 ex, yshift = 0 ex] {};
\node (i1) [draw=none,left of=c1,minimum width=1.5ex,rounded corners=0.25mm, minimum height = 6ex, xshift = 3 ex, yshift = 1.5 ex] {\scriptsize $I_s$};
\node (i2) [draw=none,left of=c1,minimum width=1.5ex,rounded corners=0.25mm, minimum height = 6ex, xshift = 3 ex, yshift = -1.5 ex] {\scriptsize $I_t$};
\draw [->,very thin] (i1) -- ($(c1.west) + (0ex,1.5ex)$);
\draw [->,very thin] (i2) -- ($(c1.west) - (0ex,1.5ex)$);
\draw [->,very thin] (c1) |- (s1);
\draw [->,very thin] ($(i1.north) - (0ex,1.5ex)$) -| ($(i1.north) + (0ex,2.5ex)$) -| (dot);
\node (loss) [draw=red!70!black,right of = dot,minimum width=2.5ex, minimum height = 2ex, xshift = -3 ex, yshift=0ex]{\scriptsize $\mathcal{L}$};
\draw [->,very thin] (dot) -- (loss);
\draw [->,very thin] ($(i2.south) + (0ex,1.5ex)$) -- ($(i2.south) - (0ex,2.5ex)$) -| (loss);
\node (img1) [draw=none,minimum width=1ex, minimum height = 2ex, xshift = 2 ex, yshift=-9.6ex]{\includegraphics[scale=0.08]{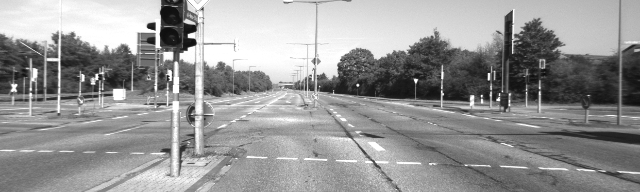}};
%\\
%
\node (img2) [draw=none, right of = img1, minimum width=1ex, minimum height = 2ex, xshift = 6 ex, yshift=0ex]{\includegraphics[scale=0.08]{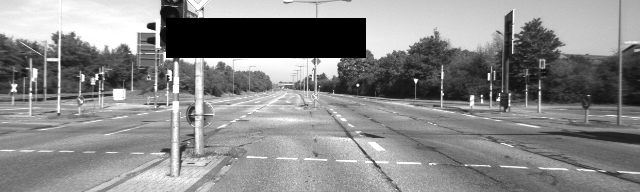}};
}};
\end{tikzpicture}
\caption{Unsupervised learning framework for \texttt{Exp$_2$}.}
\label{fig:exp2}
\end{figure} 
\subsection{Unsupervised Mask Prediction (\texttt{Exp$_2$})}
This experiment consists of two ordinary grayscale images: source ($I_s$) and target ($I_t$). From the the image $I_t$, a significantly large rectangular region is cropped out and filled with zeros. The end goal of the experiment is to predict a mask $M_s$ which depicts similar and dissimilar regions between $I_s$ and $I_t$. To achieve this, a neural network is trained in an unsupervised manner. For the learning process, we generate a dataset of $1500$ images with $1000+500$ train-test split. The dataset is generated by randomly cropping a rectangular region from $I_s$ and filling th cropped region with zeros. The obtained image is referred as $I_t$. The cropping region has a size of $40\times 200$.
\par
Fig.~\ref{fig:exp2} shows the the unsupervised training framework for this task. We use UNet \cite{unet} network architecture. The network is trained to predict $M_s$ such that $I_s * M_s \to I_t$. This experiment is essentially a binary segmentation task, therefore, we adopt intersection-over-union metric (IoU) which is widely used to quantify the segmentation performance. In order to best evaluate different training losses, we provide IoU scores by thresholding the predicted mask at different confidence levels. It is done in order to examine the network's capability to push the feature embeddings of similar and dissimilar regions significantly apart. In other words, a perfectly trained network will exhibit same IoU scores at various threshold levels. From the Table-\ref{tab:unified} under \texttt{Exp$_2$}, it is clear that the proposed $\mathcal{L}_{LMI}$ formulation shows consistent and better performance over the other loss functions across various thresholding levels.
\par
Fig.~\ref{exp2} shows a few qualitative results corresponding to this experiment. The results visualized are thresholded at $0.10$ (IoU$_{.10}$). It can be seen that $\mathcal{L}_1$ and $\mathcal{L}_{SSIM}$ performs worst whereas $\mathcal{L}_2$ and $\mathcal{L}_{LMI}$ performs marginally equally. The visual observation can also be verified quantitatively from the Table~\ref{tab:unified}, under \texttt{Exp$_2$}. In the quantitative results, $\mathcal{L}_1$ and $\mathcal{L}_{SSIM}$ performs worst in increasing order, as verified visually. On the other hand, $\mathcal{L}_2$ and $\mathcal{L}_{LMI}$ have only marginal difference, which can be also be verified visually by zooming in the results and examining the boundary of white region.
\subsection{Unsupervised Depth Estimation (\texttt{Exp$_3$})}
Depth estimation has been a long standing task in front of the computer vision community. This task has worldwide industrial importance because depth perception is a must for autonomous robotics and vehicles. Due to the advancements in the supervised learning techniques, researchers have developed several methods to predict depth using neural networks by employing supervised learning techniques. However, obtaining accurate groundtruth for supervised learning of this task is extremely challenging because it requires very expensive measurement instruments such as LIDARs. Hence, this task has gained considerable amount of attention in the recent years in order to develop unsupervised learning frameworks for this task. The existing methods make use of several loss functions in order to learn the depth effectively. 
\par
In this experiment, We demonstrate the learning of a neural network for the task of depth estimation using stereo images in an unsupervised manner. We form the seminal work for this task \cite{godard2017unsupervised} as our basis. Our aim is not to show improvements in the datasets, instead we emphasis that how the DeepMI framework can be easily integrated for these real world applications. Hence, instead of a bigger dataset, we use $87$ grayscale rectified stereo images from KITTI \cite{kitti} sequence-$113$. We select this sequence because it is quite difficult sequence from the perspective of this task.
\par
The framework to carryout the experiment is shown in Fig.~\ref{fig:exp3}. It must be noticed that instead of the three different losses as in \cite{godard2017unsupervised}, we only use one loss for the evaluation. This is done in order to demonstrate that $\mathcal{L}_{LMI}$ provides stronger gradients and alone can lead to improved results. We report MAE between the predicted depth and the groundtruth measurements. From the Table-\ref{tab:unified} under \texttt{Exp$_3$}, one can notice that the neural network trained using $\mathcal{L}_{LMI}$ outperforms other five variants by large margin. For the case of $\mathcal{L}_1, \mathcal{L}_2$, the \texttt{base\_lr} is lowered to $0.00001$ to prevent gradient explosion.
\begin{table}[!t]
\centering
\caption{ Quantitative analysis}
\label{tab:unified}

\arrayrulecolor{white!0!black}
\setlength{\arrayrulewidth}{0.15ex}
\setlength{\tabcolsep}{0.4pt}

%\footnotesize
%\scriptsize

\begin{tabular}{c c c c c c c c c}

\hline
\multirow{2}{*}{\makecell{$\mathcal{L}$}} & \multicolumn{2}{c}{\makecell{\texttt{Exp$_1$}}} & \multicolumn{5}{c}{\makecell{\texttt{Exp$_2$}}} & 
\multicolumn{1}{c}{\makecell{\texttt{Exp$_3$}}} \\ 
\cmidrule(lr){2-3}
\cmidrule(lr){4-8}
\cmidrule(lr){9-9}

& \makecell{MAE$_{tx}$} & \makecell{MAE$_{\theta}$} & \makecell{IoU$_{.05}$} & \makecell{IoU$_{.10}$} & \makecell{IoU$_{.20}$} & \makecell{IoU$_{.40}$} & \makecell{IoU$_{.50}$} & \makecell{MAE($m$)} \\  \cmidrule(lr){1-1}
\cmidrule(lr){2-3}
\cmidrule(lr){4-8}
\cmidrule(lr){9-9}

$ \mathcal{L}_1$ &  $1.52$ &  $3.44$ &  $.48$ &  $.61$&  $.65$ &  $.68$ &  $.68$ &  $105.46$ \\
$\mathcal{L}_2$ &  $1.93$ &  $3.65$ &  $.66$ &  $.68$&  $.68$ &  $.68$ &  $.68$ &  $39.78$ \\
$\mathcal{L}_{SSIM}$ &  $1.25$ &  $3.95$ &  $.50$ &  $.52$&  $.68$ &  $.69$ &  $\textcolor{blue}{.81}$ &  $.46$ \\ 
 $\mathcal{L}_{SSIM} + \mathcal{L}_{2}$ &  $-$ &  $-$ &  $-$ &  $-$&  $-$ &  $-$ &  $-$ &  $.64$ \\ 

\cmidrule(lr){1-1}
\cmidrule(lr){2-3}
\cmidrule(lr){4-8}
\cmidrule(lr){9-9}

%$\mathcal{L}_{MI}$ &  $1.24$ &  $3.58$ &  $.68$ &  $.68$&  $.68$ &  $.68$ &  $.68$ &  $.60$ \\ 

$\mathcal{L}_{LMI}$ &  $\textcolor{blue}{1.04}$ &  $\textcolor{blue}{3.18}$ &  $\textcolor{blue}{.70}$&  $\textcolor{blue}{.70}$ &  $\textcolor{blue}{.71}$ &  $\textcolor{blue}{.80}$ & $\textcolor{blue}{.81}$ & $\textcolor{blue}{.27}$ \\ \hline
\end{tabular}
\end{table}
\par
Fig.~\ref{exp3} shows a few qualitative results for this experiment. From the figure, it can be noticed that, both the $\mathcal{L}_1$ and $\mathcal{L}_{2}$ perform poorly, whereas $\mathcal{L}_{SSIM}$ performs quite better than them. This indicates the reason behind recent adaptation of $\mathcal{L}_{SSIM}$ over $\mathcal{L}_1$ and $\mathcal{L}_{2}$ for image matching purpose.
\begin{figure*}[t]
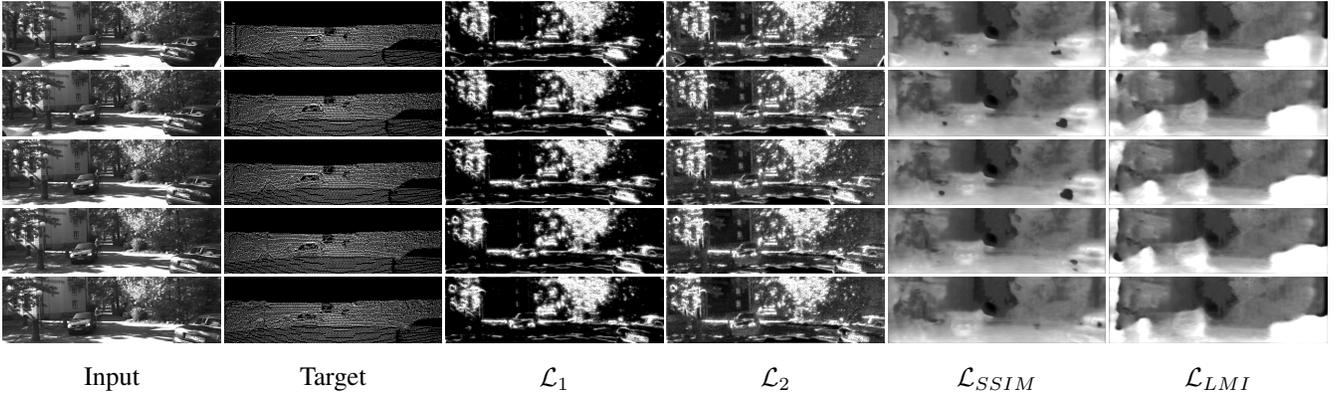

\centering
\begin{tikzpicture}
\FPeval{xoffset}{18.6}
\FPeval{yoffset}{5.8}
\FPeval{imscale}{0.128}

\foreach \i in {0,1,...,4}
{
\node (bai\i) [draw=none,rounded corners=0.6mm, xshift=0 * \xoffset ex, yshift=-(\i-1)*\yoffset ex]{\includegraphics[scale= \imscale]{de_i_l1_\i.png}};
}
\foreach \i in {0,1,...,4}
{
\node (bat\i) [draw=none,rounded corners=0.6mm, xshift=1 * \xoffset ex, yshift=-(\i-1)*\yoffset ex]{\includegraphics[scale= \imscale]{de_t_l1_\i.png}};
}
\foreach \i in {0,1,...,4}
{
\node (bat\i) [draw=none,rounded corners=0.6mm, xshift=1 * \xoffset ex, yshift=-(\i-1)*\yoffset ex]{\includegraphics[scale= \imscale]{de_d_l1_\i.png}};
}
\foreach \i in {0,1,...,4}
{
\node (bat\i) [draw=none,rounded corners=0.6mm, xshift=2 * \xoffset ex, yshift=-(\i-1)*\yoffset ex]{\includegraphics[scale= \imscale]{de_p_l1_\i.png}};
}
\foreach \i in {0,1,...,4}
{
\node (bat\i) [draw=none,rounded corners=0.6mm, xshift=3 * \xoffset ex, yshift=-(\i-1)*\yoffset ex]{\includegraphics[scale= \imscale]{de_p_l2_\i.png}};
}
\foreach \i in {0,1,...,4}
{
\node (bat\i) [draw=none,rounded corners=0.6mm, xshift=4 * \xoffset ex, yshift=-(\i-1)*\yoffset ex]{\includegraphics[scale= \imscale]{de_p_ssim_\i.png}};
}
\foreach \i in {0,1,...,4}
{
\node (bat\i) [draw=none,rounded corners=0.6mm, xshift=5 * \xoffset ex, yshift=-(\i-1)*\yoffset ex]{\includegraphics[scale= \imscale]{de_p_mi_\i.png}};
}
\FPeval{i}{5}
\node (bat\i) [draw=none,rounded corners=0.6mm, xshift=0 * \xoffset ex, yshift=-(\i-1)*\yoffset ex]{Input};
\FPeval{i}{5}
\node (bat\i) [draw=none,rounded corners=0.6mm, xshift=1 * \xoffset ex, yshift=-(\i-1)*\yoffset ex]{Target};
\FPeval{i}{5}
\node (bat\i) [draw=none,rounded corners=0.6mm, xshift=2 * \xoffset ex, yshift=-(\i-1)*\yoffset ex]{$\mathcal{L}_1$};
\FPeval{i}{5}
\node (bat\i) [draw=none,rounded corners=0.6mm, xshift=3 * \xoffset ex, yshift=-(\i-1)*\yoffset ex]{$\mathcal{L}_2$};
\FPeval{i}{5}
\node (bat\i) [draw=none,rounded corners=0.6mm, xshift=4 * \xoffset ex, yshift=-(\i-1)*\yoffset ex]{$\mathcal{L}_{SSIM}$};
\FPeval{i}{5}
\node (bat\i) [draw=none,rounded corners=0.6mm, xshift=5 * \xoffset ex, yshift=-(\i-1)*\yoffset ex]{$\mathcal{L}_{LMI}$};
\end{tikzpicture}
\caption{Qualitative results of \texttt{Exp$_3$}. More brighter the depth map, more closer is an object}.
\label{exp3}
\end{figure*}
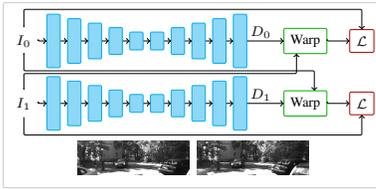
\begin{figure}[!t]
\centering

\colorlet{conv}{white!60!cyan}
\colorlet{concat}{white!60!orange}
\begin{tikzpicture}
\node (outer) [draw=white!80!black,rounded corners=0.25mm,scale = 0.75]{
\tikz{
\node (s1) [draw=cyan, fill=conv, minimum width=1.5ex,rounded corners=0.25mm, minimum height = 6ex, xshift = -4 ex, yshift = 0 ex] {};
\node (s2) [draw=cyan, fill=conv,right of = s1,rounded corners=0.25mm,minimum width=1.5ex, minimum height = 5ex, xshift = -4ex] {};
\node (s3) [draw=cyan, fill=conv,right of = s2,rounded corners=0.25mm,minimum width=1.5ex, minimum height = 4ex, xshift = -4ex] {};
\node (s4) [draw=cyan, fill=conv,right of = s3,rounded corners=0.25mm,minimum width=1.5ex, minimum height = 3ex, xshift = -4ex] {};
\node (s5) [draw=cyan, fill=conv,right of = s4,rounded corners=0.25mm,minimum width=1.5ex, minimum height = 2ex, xshift = -4ex] {};
\node (s5d) [draw=cyan, fill=conv,right of = s5,rounded corners=0.25mm,minimum width=1.5ex, minimum height = 2ex, xshift = -4ex] {};
\node (s4d) [draw=cyan, fill=conv,right of = s5d,rounded corners=0.25mm,minimum width=1.5ex, minimum height = 3ex, xshift = -4ex] {};
\node (s3d) [draw=cyan, fill=conv,right of = s4d,rounded corners=0.25mm,minimum width=1.5ex, minimum height = 4ex, xshift = -4ex] {};
\node (s2d) [draw=cyan, fill=conv,right of = s3d,rounded corners=0.25mm,minimum width=1.5ex, minimum height = 5ex, xshift = -4ex] {};
\node (s1d) [draw=cyan, fill=conv,right of=s2d, minimum width=1.5ex,rounded corners=0.25mm, minimum height = 6ex, xshift = -4 ex, yshift = 0 ex] {};
\draw [->,very thin] (s1) -- (s2);
\draw [->,very thin] (s2) -- (s3);
\draw [->,very thin] (s3) -- (s4);
\draw [->,very thin] (s4) -- (s5);
\draw [->,very thin] (s5) -- (s5d);
\draw [->,very thin] (s5d) -- (s4d);
\draw [->,very thin] (s4d) -- (s3d);
\draw [->,very thin] (s3d) -- (s2d);
\draw [->,very thin] (s2d) -- (s1d);
\node (warp) [draw=green!70!black,right of = s1d,minimum width=2.5ex, minimum height = 2ex, xshift = 1 ex, yshift=0ex]{\scriptsize Warp};
\draw [->,very thin] (s1d) -- (warp) node[xshift=-5ex, yshift=1ex]{\scriptsize $D_0$};
\node (loss) [draw=red!70!black,right of = warp,minimum width=2.5ex, minimum height = 2ex, xshift = 0 ex, yshift=0ex]{\scriptsize $\mathcal{L}$};
\draw [->,very thin] (warp) |- (loss);
\node (s12) [draw=cyan, fill=conv, minimum width=1.5ex,rounded corners=0.25mm, minimum height = 6ex, xshift = -4 ex, yshift = -7ex] {};
\node (s22) [draw=cyan, fill=conv,right of = s1,rounded corners=0.25mm,minimum width=1.5ex, minimum height = 5ex, xshift = -4ex, yshift = -7ex] {};
\node (s32) [draw=cyan, fill=conv,right of = s2,rounded corners=0.25mm,minimum width=1.5ex, minimum height = 4ex, xshift = -4ex, yshift = -7ex] {};
\node (s42) [draw=cyan, fill=conv,right of = s3,rounded corners=0.25mm,minimum width=1.5ex, minimum height = 3ex, xshift = -4ex, yshift = -7ex] {};
\node (s52) [draw=cyan, fill=conv,right of = s4,rounded corners=0.25mm,minimum width=1.5ex, minimum height = 2ex, xshift = -4ex, yshift = -7ex] {};
\node (s52d) [draw=cyan, fill=conv,right of = s52,rounded corners=0.25mm,minimum width=1.5ex, minimum height = 2ex, xshift = -4ex, yshift = 0ex] {};
\node (s42d) [draw=cyan, fill=conv,right of = s52d,rounded corners=0.25mm,minimum width=1.5ex, minimum height = 3ex, xshift = -4ex, yshift = 0ex] {};
\node (s32d) [draw=cyan, fill=conv,right of = s42d,rounded corners=0.25mm,minimum width=1.5ex, minimum height = 4ex, xshift = -4ex, yshift = 0ex] {};
\node (s22d) [draw=cyan, fill=conv,right of = s32d,rounded corners=0.25mm,minimum width=1.5ex, minimum height = 5ex, xshift = -4ex, yshift = 0ex] {};
\node (s12d) [draw=cyan, fill=conv,right of=s22d, minimum width=1.5ex,rounded corners=0.25mm, minimum height = 6ex, xshift = -4 ex, yshift = 0ex] {};
\draw [->,very thin] (s12) -- (s22);
\draw [->,very thin] (s22) -- (s32);
\draw [->,very thin] (s32) -- (s42);
\draw [->,very thin] (s42) -- (s52);
\draw [->,very thin] (s52) -- (s52d);
\draw [->,very thin] (s52d) -- (s42d);
\draw [->,very thin] (s42d) -- (s32d);
\draw [->,very thin] (s32d) -- (s22d);
\draw [->,very thin] (s22d) -- (s12d);
\node (warp2) [draw=green!70!black,right of = s12d,minimum width=2.5ex, minimum height = 2ex, xshift = 1 ex, yshift=0ex]{\scriptsize Warp};
\draw [->,very thin] (s12d) -- (warp2) node[xshift=-5ex, yshift=1ex]{\scriptsize $D_1$};
\node (loss2) [draw=red!70!black,right of = warp2,minimum width=2.5ex, minimum height = 2ex, xshift = 0 ex, yshift=0ex]{\scriptsize $\mathcal{L}$};
\draw [->,very thin] (warp2) |- (loss2);
\node (i1) [draw=none,left of=s1,minimum width=1.5ex,rounded corners=0.25mm, minimum height = 6ex, xshift = 3 ex, yshift = 0 ex] {\scriptsize $I_0$};
\node (i2) [draw=none,left of=s12,minimum width=1.5ex,rounded corners=0.25mm, minimum height = 6ex, xshift = 3 ex, yshift = 0 ex] {\scriptsize $I_1$};
\draw [->,very thin] (i1) |- (s1);
\draw [->,very thin] (i2) |- (s12);
\draw [->,very thin] ($(i1.south) + (0ex,1.5ex)$) -| ($(i1.south) - (0ex,0.3ex)$) -| ($(warp2.north) + (1ex,0ex)$);
\draw [->,very thin] ($(i2.north) - (0ex,1.5ex)$) -| ($(i2.north) + (0ex,0.3ex)$) -| ($(warp.south) - (1ex,0ex)$);
\draw [->,very thin] ($(i1.north) - (0ex,1.5ex)$) -| ($(i1.north) + (0ex,0.5ex)$) -| (loss);
\draw [->,very thin] ($(i2.south) + (0ex,1.5ex)$) -- ($(i2.south) - (0ex,0.5ex)$) -| (loss2);
\node (img1) [draw=none,minimum width=1ex, minimum height = 2ex, xshift = 5 ex, yshift=-13.1ex]{\includegraphics[scale=0.045]{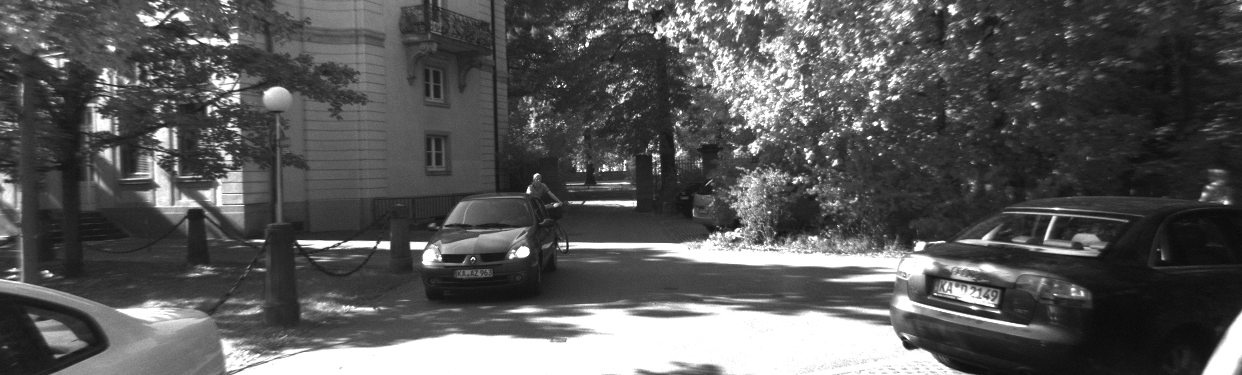}};
%\\
%
\node (img2) [draw=none, right of = img1, minimum width=1ex, minimum height = 2ex, xshift = 7 ex, yshift=0ex]{\includegraphics[scale=0.045]{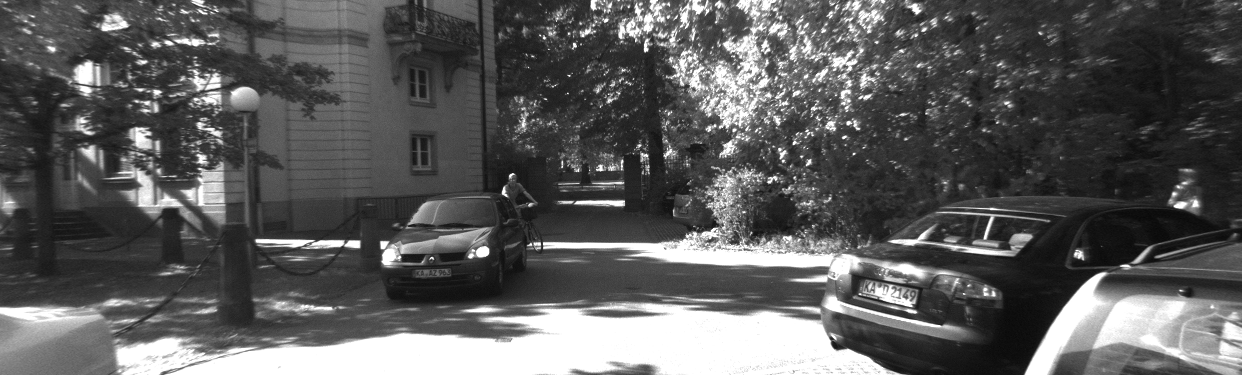}};
}};
\end{tikzpicture}
\caption{Unsupervised learning framework for \texttt{Exp$_3$}. Warp~\cite{godard2017unsupervised}.}
\label{fig:exp3}
\end{figure} 
\par
Further, it can be noticed that, visually the results of $\mathcal{L}_{LMI}$ are the most pleasing as well as consistent among all. As a comparison between the $\mathcal{L}_{SSIM}$ and $\mathcal{L}_{LMI}$, we can see that the depth estimations of the former contains texture copy \cite{godard2017unsupervised} artifacts, whereas these are absent in latter. It can also be verified by taking a closer look at the car in the bottom right of the image. It can be seen that for the case of $\mathcal{L}_{SSIM}$, the depth map of the car has holes near edges and also contains severe texture copy artifacts near the number plate and other body area of the car. These artifacts, on the other hand are not present in the case of $\mathcal{L}_{LMI}$. This shows the clear effectiveness of $\mathcal{L}_{LMI}$ loss and the DeepMI framework.

\subsection{Effect of number of bins $N$}
\label{sec:exphparams}
Table-\ref{tab:effectN} shows the effect of the hyperparameter $N$ on each of the three tasks. For \texttt{Exp$_1$}, it can be noticed that the performance for all the four values of $N$ is same. It has to be the case because the images are binary in this task. For \texttt{Exp$_2$}, there is considerable drop in the performance for $N=3$ for IoU$_{0.05}$ (highlighted in \textcolor{red}{red}). It is because, the images in this case have multiple grayscale levels and details about which can not be efficiently captured. The same is the case for \texttt{Exp$_3$} when $N=3$. Overall, we can see that $\mathcal{L}_{LMI}$ has significant advantages over other loss functions. Through our experimentations, it sufficient to keep the value of $N \leq 25$ for a signal having dynamic range $\in[0,255]$.
\subsection{A unified discussion on the experiments}
\label{sec:compminmi}
From the Table-\ref{tab:unified}, it can be noticed that the information theory based measures are much more consistent as compared to the other losses. Also, it can be noticed that for the \texttt{Exp$_3$}, the $\mathcal{L}_{LMI}$ proves to be better over the case when $\mathcal{L}_{SSIM}$ and $\mathcal{L}_2$ are used together. Also, it is noticeable that $\mathcal{L}_{LMI}$ shows consistent and best scores amongst all variants of losses. In the experiments, our intention has not been to outweigh the existing losses, instead to mark the potential of information theory based methods in deep learning for real world applications.
\begin{table}[t]
\centering
\caption{Effect of bin size $N$ on $\mathcal{L}_{LMI}$}
\label{tab:effectN}

\arrayrulecolor{white!0!black}
\setlength{\arrayrulewidth}{0.15ex}
\setlength{\tabcolsep}{2.5pt}

%\footnotesize
%\scriptsize

\begin{tabular}{c c c c c c c c c}
\hline
\multirow{2}{*}{\makecell{$N$}} & \multicolumn{2}{c}{\makecell{\texttt{Exp$_1$}}} & \multicolumn{5}{c}{\makecell{\texttt{Exp$_2$}}} & 
\multicolumn{1}{c}{\makecell{\texttt{Exp$_3$}}} \\ 
\cmidrule(lr){2-3}
\cmidrule(lr){4-8}
\cmidrule(lr){9-9}

& \makecell{MAE$_{tx}$} & \makecell{MAE$_{\theta}$} & \makecell{IoU$_{.05}$} & \makecell{IoU$_{.10}$} & \makecell{IoU$_{.20}$} & \makecell{IoU$_{.40}$} & \makecell{IoU$_{.50}$} & \makecell{MAE($m$)} \\  \cmidrule(lr){1-1}
\cmidrule(lr){2-3}
\cmidrule(lr){4-8}
\cmidrule(lr){9-9}

$3$ &  $1.04$ &  $3.18$ &  \textcolor{red}{$.41$} &  $.51$&  $.52$ &  $.50$ &  \textcolor{blue}{$.81$} &  $.41$\\ 
$11$ &  $0.89$ &  \textcolor{blue}{$2.69$} &  $.50$ &  $.59$&  $.61$ &  $.66$ &  \textcolor{blue}{$.81$} &  \textcolor{blue}{$.27$}\\
$15$ &  \textcolor{blue}{$0.78$} &  $2.89$ &  \textcolor{blue}{$.66$} &  \textcolor{blue}{$.67$}&  \textcolor{blue}{$.68$} &  $.69$ &  \textcolor{blue}{$.81$} &  $.28$\\ 
$25$ &  $0.99$ &  $4.36$ &  $.47$ &  $.48$&  $.58$ &  \textcolor{blue}{$.79$} &  $.79$ &  $.29$\\ \hline
\end{tabular}
\end{table}
\section{Conclusion}
\label{sec:conc}
In this paper, we proposed an end-to-end differentiable framework ``DeepMI'' and a novel similarity metric $\mathcal{LMI}$ to train deep neural networks. The metric is mutual information ($\mathcal{MI}$) inspired and cops up with the difficulty to be able to train a deep neural network using the ($\mathcal{MI}$) expression. The metric is based on probability density functions which makes it signal agnostic. The density functions are discrete in nature for real world signals (images, time series) and can not support backpropagation. Therefore, a fuzzification strategy to support smooth backward gradient flow through the density functions is also developed. We show that the neural network trained using $\mathcal{L}_{MI}$ metric, outperforms their counterparts which are trained using $\mathcal{L}_1, \mathcal{L}_2, \mathcal{L}_{SSIM}$. Additionally, we also show that it can be easily integrated for real world applications.
\par
The DeepMI framework can be thought of as an effort to club the deep learning and mutual information together for real world applications. Through this work, we believe that, the learning based methods in several areas such as autonomous vehicles, robotic vision, speech / audio can be greatly benefitted in terms of the performance. The extensive experimental study in this work, can be used as a ground to develop extensions to the DeepMI framework in order to further improve the overall algorithmic performance. We also believe that inclusion of DeepMI into deep learning frameworks shall open door to new applications.

\medskip
\small

%\nocite{*}
\bibliographystyle{ieeetr}
\bibliography{bibfile}

\end{document}